\documentclass{article}
\usepackage{spconf,amsmath,graphicx}
\usepackage{amssymb,amsmath,epsfig, bm}
\ninept

\title{Knowledge Distillation for Small-footprint Highway Networks}
\name{ Liang Lu$^1$ \qquad  Michelle Guo$^{\dag 2}$ \qquad Steve Renals$^3$\thanks{$^\dag$The work was partially done when M. Guo was visiting The University of Edinburgh under the Stanford Bing Overseas Studies Programme.}}

\address{ $^1$TTI-Chicago $\quad$ $^2$Stanford University $\quad$ $^3$The University of Edinburgh \\
{\small \tt llu@ttic.edu $\quad$ mguo95@stanford.edu  $\quad$ s.renals@ed.ac.uk}}

\begin{document}
%
\maketitle
\begin{abstract}
Deep learning has significantly advanced state-of-the-art of speech recognition in the past few years. However, compared to conventional Gaussian mixture acoustic models, neural network models are usually much larger, and are therefore not very deployable in embedded  devices. Previously, we investigated a compact highway deep neural network (HDNN) for acoustic modelling, which is a type of depth-gated feedforward neural network. We have shown that HDNN-based acoustic models can achieve comparable recognition accuracy with much smaller number of model parameters compared to plain deep neural network (DNN) acoustic models. In this paper, we push the boundary further by leveraging on the knowledge distillation technique that is also known as {\it teacher-student} training, i.e., we train the compact HDNN model with the supervision of a high accuracy cumbersome model. Furthermore, we also investigate sequence training and adaptation in the context of teacher-student training. Our experiments were performed on the AMI meeting speech recognition corpus. With this technique, we significantly improved the recognition accuracy of the HDNN acoustic model with less than 0.8 million parameters,  and narrowed the gap between this model and the plain DNN with 30 million parameters.

\end{abstract}

\section{Introduction}
\label{sec:intro}

Recent years have witnessed wide applications of speech technology in embedded devices like mobile phones, thanks to deep learning that has significantly advanced state-of-the-art in this area. For scenarios that internet connections are unavailable or for privacy concerns, it is desirable that speech recognisers can run locally in such kind of resource constrained platforms. However, state-of-the-art neural network models are either computationally expensive or consume large amount of memory, and are therefore unsuitable for this purpose. Recently, there have been a number of works on small footprint acoustic models to address this problem such as using low-rank matrices~\cite{xue2013restructuring, sainath2013low}, structured linear layers~\cite{le2013fastfood, sindhwani2015structured, moczulski2015acdc}, and the use of low rank displacement of structured matrices~\cite{sindhwani2015structured}. Instead of manipulating the model parameters, another approach is based on the {\it teacher-student} architecture~\cite{li2014learning, ba2014deep, romero15_fitnet}, which is also known as model compression~\cite{bucilu2006model} or knowledge distillation~\cite{hinton2015distilling}. In this approach, the {\it teacher} may be a large-size network or an ensemble of several different models, which is used to predict the soft targets for training the {\it student} model that is much smaller. As pointed out in~\cite{hinton2015distilling}, the soft targets provided by the teacher encode the generalisation power of the teacher, and the student model trained using these pseudo labels is observed to perform better than the same model trained independently using the ground truth labels~\cite{hinton2015distilling}. 

Previously, we studied a compact acoustic model using highway deep neural network (HDNN) for resource constrained speech recognition~\cite{llu2016a}. HDNN is a type of network with shortcut connections between hidden layers~\cite{srivastava2015training}. Compared to the plain networks with skip connections, HDNNs are equipped with two gate functions --  {\it transform} and {\it carry} gate -- to control and facilitate the information flow over all the whole network.  In particular, the transform gate is used to scale the output of a hidden layer and the carry gate is used to pass through the input directly after elementwise rescaling. The gate functions are the key to train very deep networks~\cite{srivastava2015training} and to speed up convergence as experimentally validated in~\cite{llu2016a}. We have shown that the gate functions can manipulate the behavior of the whole neural networks in sequence training and adaptation~\cite{lu2016sequence}. With the gate functions, we can train much thinner and deeper networks with much smaller number of model parameters, which can achieve comparable recognition accuracy compared to much larger plain DNNs.  

In this paper, we investigate teacher-student training to further improve the accuracy of the small-footprint HDNN acoustic model. In particular, we use a large size plain DNN acoustic model to provide soft labels for training the student HDNN model. As mentioned before, there have been a number of work on teacher-student training for speech recognition~\cite{li2014learning, ba2014deep, hinton2015distilling}. The one that is closest to our study is~\cite{li2014learning}. However, the student model investigated in this paper is much smaller due to the highway architecture. In addition, we present further analysis and experimental study on hybrid loss functions that interpolate the cross-entropy and teacher-student costs, the use of temperature to smooth the soft labels as well as sequence training and adaptation results in this context. 

\section{Highway Deep Neural Networks}
\label{sec:hdnn}

In this work, we focus on feed-forward neural networks -- also known as DNNs -- as target student models. Although long short-term memory based recurrent neural networks (LSTM-RNNs) and convolutional neural networks (CNNs) may obtain higher recognition accuracy compared to DNNs~\cite{sak2014long, abdel2014convolutional}, they are more suitable for teacher models because they are usually computationally more expensive for applications on resource constrained platforms. A plain DNN with $L$ hidden layers may be represented as
\begin{align}
\bm h_t^{(1)} &= \sigma(\bm x_t, \theta_1) \\
\bm h_t^{(l)} &= \sigma(\bm h_t^{(l-1)}, \theta_l), \quad \text{for} \quad l=2,\ldots, L \\
\label{eq:sm}
\bm y_t &= g(\bm h_t^{(L)}, \varphi)
\end{align}
where $\bm x_t$ is an input vector to the network at the time step $t$; $\sigma(\bm h_t^{(l-1)}, \theta_l)$ denotes the transformation of the input $\bm h_t^{(l-1)}$ with the parameter $\theta_l$ followed by a nonlinear activation function, e.g., {\tt sigmoid}; $g(\cdot, \varphi)$ is the output function that is parameterised by $\varphi$ in the output layer, which usually uses the softmax to obtain the posterior probability of each class given the input feature.    

Highway deep neural networks~\cite{srivastava2015training} augment the feature extractor with gate functions, in which the hidden layer may be represented as
\begin{align}
\label{eq:hw}
{\bm h}_t^{(l)} &= \sigma({\bm h}_t^{(l-1)}, \theta_l)\circ T({\bm h}_t^{(l-1)}, {\bm W}_T) \nonumber \\
& \qquad \quad + {\bm h}_t^{(l-1)}\circ C({\bm h}_t^{(l-1)}, {\bm W}_c), 
\end{align}
where $T(\cdot)$ is the {\it transform} gate that scales the original hidden activations; $C(\cdot)$ is the {\it carry} gate, which scales the input before passing it directly to the next hidden layer; $\circ$ denotes elementwise multiplication; The outputs of $T(\cdot)$ and $C(\cdot)$ are constrained to be within $[0, 1]$, and we use the sigmoid function for both gates that are parameterised by $\mathbf{W}_T$ and $\mathbf{W}_c$ respectively. Following our previous work~\cite{srivastava2015training}, we tie the parameters in the gate functions across all the hidden layers, which can significantly save model parameters. In this work, we do not use any bias vector in the two gate functions. As pointed out in~\cite{lu2016sequence}, to speed up the training, we can pack the matrices as
\begin{align} 
\tilde{\bm W}_l = \left[ \bm W_l^\top, \bm W_T^\top, \bm W_c^\top \right]^\top,
\end{align}
where $\bm W_l^\top$ is the weight matrix in the $l$-th layer, and we then compute $\tilde{\bm W}_l\bm h_{l-1}$ once for all. By this trick, we can leverage on the power of GPUs on computing large matrix-matrix multiplications efficiently in the minibatch mode.

\section{Model Training}
\label{sec:train}

\subsection{Cross-Entropy Training}
\label{sec:ce}

The most common criterion to train neural networks for classification is cross-entropy (CE), which defines the loss function as 
\begin{align}
\label{eq:ce}
\mathcal{L}^{(CE)}(\theta) = - \sum_j \hat{y}_{jt} \log y_{jt}, 
\end{align}
where $j$ is the index of the hidden Markov model (HMM) state; $\bm y_t$ is the output of the nerual network as Eq. \eqref{eq:sm}, while $\hat{\bm y}_t$ denotes the ground truth label that is a one-hot vector. Note that, the loss function is defined with one training utterance here for the simplicity of notation. Supposing that $ \hat{y}_{jt} = \delta_{ij}$, where $\delta_{ij}$ is the Kronecker delta function and $i$ is the ground truth class at the time step $t$,  the CE loss becomes 
\begin{align}
\label{eq:ce}
\mathcal{L}^{(CE)}(\theta) = - \log y_{it}.
\end{align}
In this case, minimising $\mathcal{L}^{(CE)}(\theta)$ is equivalently to minimise the negative log posterior probability of the correct class, and it is equal to maximising the probability $y_{it}$, while the posterior probabilities of other classes are ignored. However, maximising $y_{it}$ will also result in minimising the posterior probabilities of other classes since they sum to one.

\subsection{Teacher-Student Training}

Instead of using the ground truth labels, the teacher-student training approach defines the loss function as
\begin{align}
\label{eq:kd}
\mathcal{L}^{(KD)}(\theta) = - \sum_j \tilde{y}_{jt} \log y_{jt},
\end{align}
where $\tilde{y}_{jt}$ is the output of the teacher model, which works as a pseudo label. As pointed out in~\cite{li2014learning}, the loss function as Eq. \eqref{eq:kd} is equivalent to minimise the Kullback-Leibler divergence between the posterior probabilities of each class from the teacher and student models. Here, $\tilde{y}_{jt}$  is no longer a one-hot vector, instead, the competing classes will have small but nonzero posterior probabilities for each training example. Hinton et al.~\cite{hinton2015distilling} suggested that the small posterior probabilities are valuable information that encodes correlations among different classes. However, their roles may be very small in the loss function as these probabilities are close to zero due to the softmax function. To address this problem, they suggested to use a large temperature to flatten the posterior distribution as
\begin{align}
y_{jt} &= \frac{\exp \left(z_{jt}/T\right)}{\sum_{i} \exp \left(z_{it}/T\right)}, \\
\bm z_t &= \bm W^{(L+1)}\bm h_t^{(L)} + \bm b^{(L+1)},
\end{align}
where $\bm W^{(L+1)}, \bm b^{(L+1)}$ are parameters in the softmax layer, and $T\in \mathbb{R}^+$ is the temperature. Following~\cite{hinton2015distilling}, we applied the same temperature to the softmax functions in both the teacher and student networks in our experiments, as only increasing the temperature in the teacher network resulted in much higher error rates in our experiments. 

One particular advantage of the teacher-student training approach is that unlabelled  data can be used easily. However, when the ground truth labels are available, it may be beneficial to incorporate the ground truth information into the loss function, which can be done by interpolating the two loss functions as
\begin{align}
\label{eq:hybrid}
\widetilde{\mathcal{L}(\theta)} = \mathcal{L}^{(KD)}(\theta) + q\mathcal{L}^{(CE)}(\theta)
\end{align}
where $q\in \mathbb{R}^+$ is the tuning parameter. We denote this as the hybrid loss, and it will be studied in the experimental section. 

\subsection{Sequence Training}

While the previous two loss functions are defined at the frame level, sequence training defines the loss at the sequence level, which usually yields significant improvement for speech recognition~\cite{kingsbury2012scalable, Vesely:IS13, su2013error}. If we denote ${\bm X}$ as the sequence of acoustic frames ${\bm X} = \{\bm x_1, \ldots, \bm x_T\}$ and $\bm Y$ as the sequence of labels, where $T$ is the length of the signal, the loss function from the state-level minimum Bayesian risk criterion(sMBR)~\cite{gibson2006hypothesis, kingsbury2009lattice} is defined as
\begin{align}
\mathcal{L}^{(sMBR)}(\theta) = \frac{\sum_{\mathcal{W} \in \Phi}p({\bm X} \mid \mathcal{W})^k P(\mathcal{W})A(\bm Y, \hat{\bm Y})}{\sum_{\mathcal{W} \in \Phi}p({\bm X} \mid \mathcal{W})^k P(\mathcal{W})},
\end{align}
where $A(\bm Y, \hat{\bm Y})$ measures the state level distance between the ground truth and predicted labels; $\Phi$ denotes the hypothesis space represented by a denominator lattice, and $\mathcal{W}$ is the word-level transcription; $k$ is the acoustic score scaling parameter. In this paper, we only focus on the sMBR criterion since it can achieve comparable or slightly better results compared to the maximum mutual information (MMI) or minimum pone error (MPE) criterion~\cite{Vesely:IS13}.

For sequence training, the acoustic model is normally firstly trained with the CE loss function, which is then fine tuned with the sequence-level loss for a few iterations. While for knowledge distillation, the model is firstly trained with the loss function as Eq. \eqref{eq:kd}. This may raise the question that if the improvement will diminish in sequence training, and we will perform experimental study to answear this question. Note that, only applying the sequence training criterion without regularisation may lead to overfitting as observed in~\cite{ Vesely:IS13, su2013error}. To address this problem, we interpolate the sMBR loss function with the CE loss~\cite{su2013error}. However, for the case of knowledge distillation, we apply the following interpolation:
\begin{align}
\label{eq:reg}
\widehat{\mathcal{L}(\theta)} = \mathcal{L}^{(sMBR)}(\theta) + p \mathcal{L}^{(KD)}(\theta), 
\end{align}
where $p\in\mathbb{R}^+$ is the smoothing parameter.

\section{Experiments}
\label{sec:exp}

\subsection{System Setup}

Our experiments were performed on the individual headset microphone (IHM) subset of the AMI meeting speech transcription corpus~\cite{renals2007recognition}.The amount of training data is around 80 hours, corresponding to roughly 28 million frames. We followed the experimental setup in~\cite{lu2016sequence}. We used 40-dimensional fMLLR adapted features vectors normalised on the per-speaker level, which were then spliced by a context window of 15 frames (i.e. $\pm7$). The number of tied HMM states is 3927. The HDNN models were trained using the CNTK toolkit~\cite{yu2014introduction}, while the results were obtained using the Kaldi decoder~\cite{povey2011kaldi}. We also used the Kaldi tookit to compute the alignment and lattices for sequence training. We set the momentum to be 0.9 after the 1st epoch for CE training, and we used the sigmoid activation for all the networks. The weights in each hidden layer of HDNNs were randomly initialised with a uniform distribution in the range of $[-0.5, 0.5]$ and the bias parameters were initialised to be $0$ for CNTK systems. We used a trigram language model for decoding. The word error rates (WERs) of the baseline systems with different model structures are shown in Table \ref{tab:base}.

\begin{table}[t]
\caption{Baseline results of DNN and HDNN systems with CE and sMBR training. The DNN systems were built using Kaldi toolkit, where the networks were pre-trained using restricted Bolzman machines. Results are shown in terms of word error rates (WERs). We use $H$ to denote the size of hidden units, and $L$ the number of layers. } \vskip 1mm
\label{tab:base}
\centering \small
\begin{tabular}{lc|cc|cc}
\hline 

\hline
  & & \multicolumn{2}{c|}{{\tt eval}} & \multicolumn{2}{c}{\tt dev}  \\
Model  & Size & CE & sMBR & CE & sMBR  \\ \hline
DNN-$H_{2048}L_{6}$ & $30 M$  & 26.8  & 24.6 & 26.0 & 24.3 \\
DNN-$H_{512}L_{10}$ & $4.6 M$ & 28.0 & 25.6 & 26.8 & 25.1\\
DNN-$H_{256}L_{10}$ & $1.7 M$ & 30.4 & 27.5 & 28.4 & 26.5\\ 
DNN-$H_{128}L_{10}$ & $0.71 M$ & 34.1 & 30.8 & 31.5 & 29.3 \\ \hline
HDNN-$H_{512}L_{10}$ & $5.1 M$ & 27.2 & 24.9 & 26.0 & 24.5 \\
HDNN-$H_{256}L_{10}$ & $1.8 M$ & 28.6 & 26.0 & 27.2 & 25.2 \\
HDNN-$H_{128}L_{10}$ & $0.74 M$ & 32.0 & 29.4 & 29.4 & 28.1 \\
HDNN-$H_{512}L_{15}$ & $6.4 M$ & 27.1 & 24.7 & 25.8 & 24.3 \\
HDNN-$H_{256}L_{15}$ & $2.1 M$ & 28.4 & 25.9 & 26.9 & 25.2 \\ \hline

 \hline
\end{tabular}
\vskip-4mm
\end{table}

\subsection{Loss Function and Temperature}

\begin{figure}[t]
\small
\centerline{\includegraphics[width=0.5\textwidth]{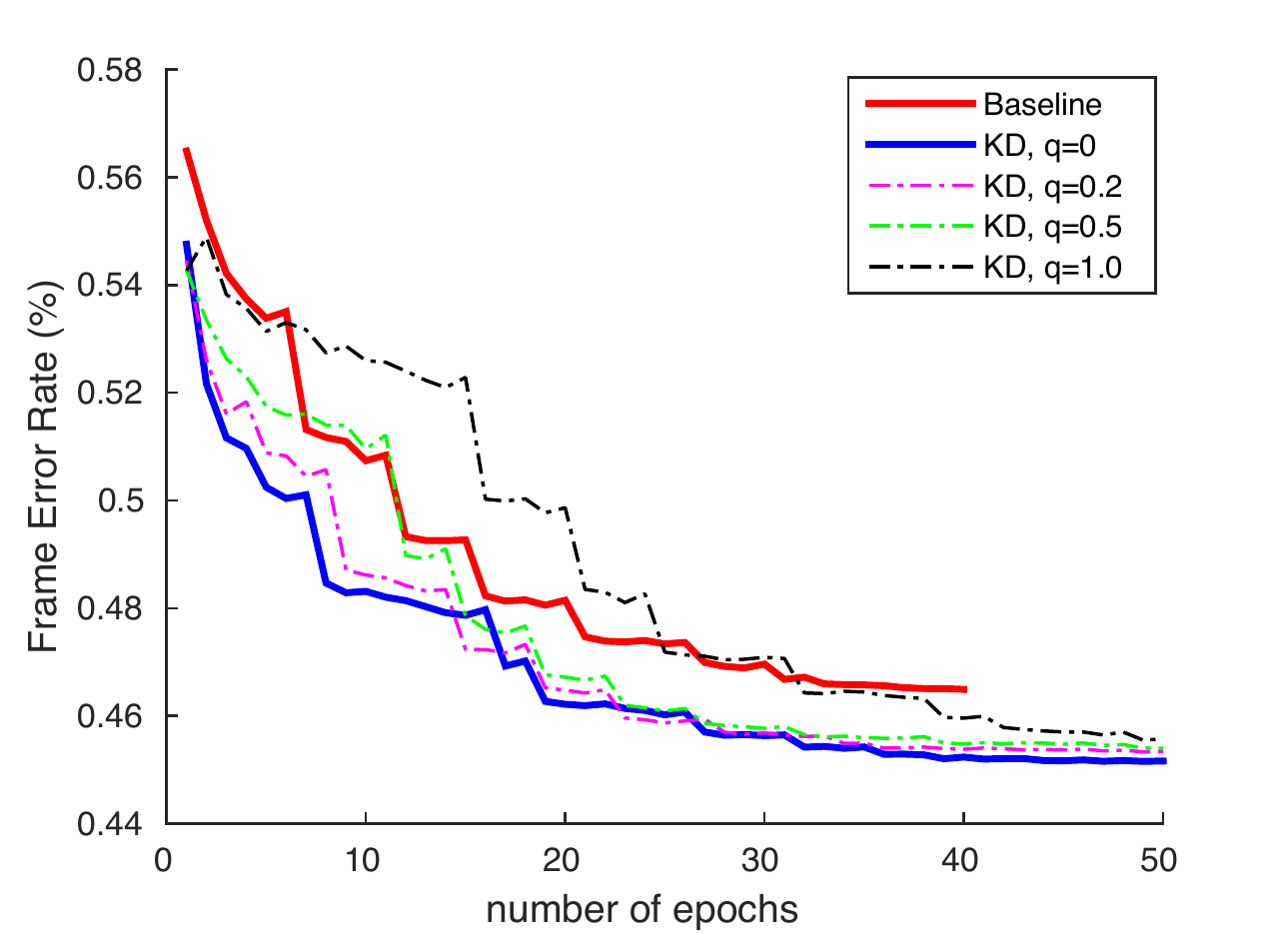}} 
\caption{Convergence curves of teacher-student training. The frame error rates were obtained from the cross validation set.  It slows down the convergence as $q$ increases. }  \vskip-2mm
\label{fig:kd}
\vskip-3mm
\end{figure}

We firstly compare the teacher-student loss function as Eq. \eqref{eq:kd} and the hybrid loss function as Eq. \eqref{eq:hybrid}. We used a CE trained plain DNN-$H_{2048}L_6$ as the teacher model, and used the HDNN-$H_{128}L_{10}$ as the student model. Figure \ref{fig:kd} shows the convergence curves when training the model with different loss functions, while Table \ref{tab:kd} shows the WERs. We observe that by teacher-student training without the ground truth labels, we can achieve significantly lower frame error rate on the cross validation set as shown in Figure \ref{fig:kd}, corresponding to moderate WER reduction (31.3\% vs. 32.0 on the {\tt eval} set). However, using the hybrid loss function as Eq. \eqref{eq:hybrid}, we do not obtain further improvement. In fact, it converges slower when $q>0$ during training as shown in Figure \ref{fig:kd}. Our interpretation is that it may be because the probabilities of uncorrected classes played a smaller role in this case, which supports the argument that they encode useful information for training the student model~\cite{hinton2015distilling}. This hypothesis encouraged us to investigate the use of a large temperature to flatten the posterior probability distribution of the labels from the teacher model. The results are also shown in Table \ref{tab:kd}. Contrary to our expectation, using large temperatures results in higher WERs. In the following experiments, we fixed $q=0$ and $T=1$. 

\begin{table}[t]
\caption{Results of teacher-student training with different loss functions and temperatures.} \vskip 1mm
\label{tab:kd}
\centering \small
\begin{tabular}{lcc|cc}
\hline 

\hline
  & & & \multicolumn{2}{c}{WER}   \\
Model  & $q$ & $T$ & {\tt eval} & {\tt dev}   \\ \hline
DNN-$H_{128}L_{10}$ & -- & -- & 34.1  & 31.5  \\ 
HDNN-$H_{128}L_{10}$ baseline & -- & -- & 32.0 & 29.9  \\ \hline
HDNN-$H_{128}L_{10}$  & 0 & 1 &  31.3 & 29.3  \\
HDNN-$H_{128}L_{10}$  & 0.2 & 1 & 31.4 & 29.5  \\
HDNN-$H_{128}L_{10}$  & 0.5 & 1& 31.3 & 29.4  \\
HDNN-$H_{128}L_{10}$  & 1.0 & 1 & 31.3 & 29.4 \\ \hline
HDNN-$H_{128}L_{10}$  & 0 & 2 & 32.3 & 29.9 \\ 
HDNN-$H_{128}L_{10}$  & 0 & 3 & 33.0 & 30.6 \\ \hline

 \hline
\end{tabular}
\vskip-4mm
\end{table}

\begin{figure}
\small
\centerline{\includegraphics[width=0.5\textwidth]{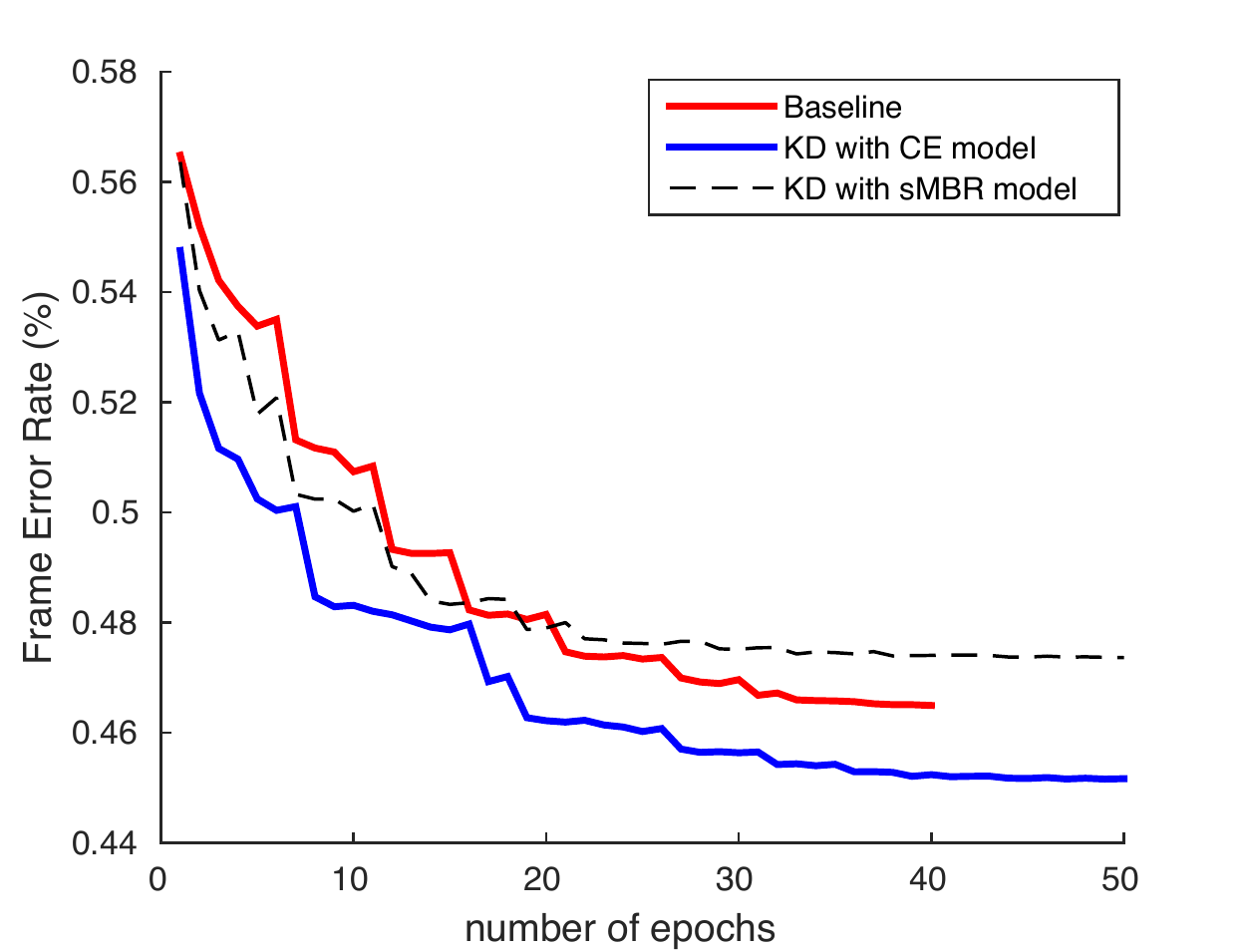}} 
\caption{Convergence curves of teacher-student training with CE or sMBR-based teacher model. }  \vskip-2mm
\label{fig:seq}
\vskip-3mm
\end{figure}

\begin{table}[t]
\caption{Results of sequence training on the {\tt eval} set for the student HDNN model. LR denotes the learning rate.} \vskip 1mm
\label{tab:seq}
\centering \small
\begin{tabular}{l|cccc}
\hline 

\hline
Teacher & LR & $p$ & $\mathcal{L}^{(KD)} \rightarrow \widehat{\mathcal{L}(\theta})$   \\ \hline
DNN-$H_{2048}L_{6}$-CE  & $1\times 10^{-5}$ & 0.2 & 31.3 $\rightarrow$ 28.4 \\ 
DNN-$H_{2048}L_{6}$-sMBR& $1\times10^{-5}$ & 0.2 & 28.8 $\rightarrow$ 28.9 \\ 
DNN-$H_{2048}L_{6}$-sMBR& $1\times10^{-5}$ & 0.5 & 28.8 $\rightarrow$ 28.0 \\  
DNN-$H_{2048}L_{6}$-sMBR& $5\times10^{-6}$ & 0.2 & 28.8 $\rightarrow$ 28.6 \\  
DNN-$H_{2048}L_{6}$-sMBR& $5\times10^{-6}$ & 0.5 & 28.8 $\rightarrow$ 28.0 \\ \hline 

 \hline
\end{tabular}
\vskip-5mm
\end{table}

\subsection{Teacher Model}

We then improved the teacher model by sMBR-based sequence training, and used this model to supervise the training of the student model. Similar to the observations in~\cite{li2014learning}, the sMBR-based teacher model can significantly improve the performance of the student model. In fact, the error rate is lower than that achieved by the student model trained independently with sMBR as shown in Table \ref{tab:seq} (28.8\% vs. 29.4\% on the {\tt eval} set). Note that, since the sequence training criterion is not to maximise the frame accuracy, training the model with this criterion normally reduces the frame accuracy, as shown explicitly by Figure 6 in~\cite{heigold2014asynchronous}. Interestingly, we observed the same pattern in the case of teacher-student training. Figure \ref{fig:seq} shows the convergence curves of using CE and sMBR based teacher models, where we see that the student model achieves much higher frame error rate on the cross validation set when supervised by sMBR-based teacher model, though the loss function as Eq. \eqref{eq:kd} is at the frame level.

We then study if the accuracy of the student model can be further improved by the sequence level criterion. Here, we set the smoothing parameter $p=0.2$ in Eq. \eqref{eq:reg} and the default learning rate to be $1\times 10^{-5}$ following our previous setup in~\cite{lu2016sequence}. Table \ref{tab:seq} shows the sequence training results of student models supervised by the CE and sMBR-based teacher models respectively. Not surprisingly, the student model supervised by the CE-based DNN model can be significantly improved by the sequence training. Notably, the WER obtained by this approach is lower compared to the model trained independently with sMBR (28.4\% vs. 29.4\% on the {\tt eval} set). However, this configuration did not work for the student model supervised by the sMBR-based teacher model. After inspection, we found that it was due to overfitting.  We then increased the value of $p$ for stronger regularisation and reduced the learning rate. Lower WERs can be obtained as the table shows, however, the improvement is less significant as the sequence level information has already been integrated into the teacher model.

\begin{table}[t]
\caption{Results of unsupervised speaker adaptation. DP denotes the number of decoding passes.} \vskip 1mm
\label{tab:adap}
\centering \small
\begin{tabular}{l|ccccc}
\hline 

\hline
 &   & & & \multicolumn{2}{c}{{\tt eval}}   \\
Model & Loss & Update &DP& SI & SD  \\ \hline
HDNN-$H_{128}L_{10}$ & $\mathcal{L}^{CE}$ & All & 2& 29.4 & 28.8 \\
HDNN-$H_{128}L_{10}$ & $\mathcal{L}^{CE}$ & Gates &2 & 29.4 & 28.7  \\
HDNN-$H_{128}L_{10}$-KD &$\mathcal{L}^{KD}$ & All  &1&  28.4 & 27.5 \\ 
HDNN-$H_{128}L_{10}$-KD &$\mathcal{L}^{KD}$ & Gates &1 & 28.4 & 27.8 \\ 
HDNN-$H_{128}L_{10}$-KD &$\mathcal{L}^{CE}$ & All  & 2& 28.4 & 27.7 \\ 
HDNN-$H_{128}L_{10}$-KD &$\mathcal{L}^{CE}$ & Gates& 2  & 28.4 & 27.1 \\ \hline

 \hline
\end{tabular}
\vskip-5mm
\end{table}

\subsection{Unsupervised Adaptation}
Our final experiments concern adaptation. Neural network acoustic models are less adaptable due to large number of unstructured model parameters compared to conventional acoustic models using Gaussian mixtures. However, a smaller model may be easier to adapt. In particular, the gate functions in HDNNs are more adaptable since they have much smaller number of model parameters, e.g., the total number of parameters in $(\bm W_T, \bm W_C)$ of an HDNN-$H_{128}L_{10}$ acoustic model is around 0.03 million. Furthermore, only updating the gate functions does not easily yield overfitting with small amount of adaptation data and pseudo labels~\cite{lu2016sequence}. We performed similar adaptation experiments for HDNN trained by the teacher-student approach. We applied the second-pass adaptation approach for the standalone HDNN model, i.e., we decoded the evaluation utterances to obtain the hard labels first, and then used these labels to adapt the model using the CE loss as Eq. \eqref{eq:ce}. However, using the teacher-student loss as Eq. \eqref{eq:kd}, only one pass decoding is required because the pseudo labels for adaptation are provided by the teacher model, which does not need the word level transcription. This is a particular advantage of the teacher-student training technique. However, note that for resource constrained application scenarios, the student model should be adapted offline, because otherwise the teacher model needs to be accessed to generate the labels. This requires another set of unlabelled speaker-dependent data for adaptation, but it is usually not expensive to collect. 
 
Since the standard AMI corpus does not have this additional set of speaker-dependent data, we only show online adaptation results. We used the teacher-student trained model from row 1 of Table \ref{tab:seq} as the speaker-independent (SI) model because its pipeline is much simpler. The baseline system used the same network as the SI model, but it was trained independently. During adaptation, we updated the SI model by 5 iterations with fixed learning rate as $2\times 10^{-4}$ per sample following our previous setup~\cite{lu2016sequence}. We also compared the CE loss as Eq. \eqref{eq:ce} and the teacher-student loss as Eq.~\eqref{eq:kd} for adaptation. Results are given in Table \ref{tab:adap}. Using the CE loss function for both SI models, only updating the gates yields slightly better results, while updating all the model parameters gives smaller improvements, possibly due to overfitting. Interestingly, this is not the case for the teacher-student loss, i.e. updating all the model parameters yields lower WER. These results may agree with the argument in~\cite{hinton2015distilling} that the soft targets can work as a regulariser and can prevent the student model from overfitting. 

\section{Conclusions}

In this paper, we investigated the teacher-student training for small-footprint acoustic models using HDNNs. We observed that the accuracy of the student acoustic model could be improved under the supervision of a high accuracy teacher model, even without additional unsupervised data. In particular, the student model supervised by a sMBR-based teacher model achieved lower WER compared to the model trained independently using the sMBR-based sequence training approach. Unsupervised speaker adaptation further improved the recognition accuracy by around 5\% relative for our model with less then 0.8 million model parameters.  However, we did not obtain improvements by using the hybrid loss function by interpolating the CE and teacher-student loss functions, and using higher temperature to smooth the pseudo labels did not help either. In the future, we shall evaluate this model on low resource conditions where the amount of training data is much smaller. 


\bibliographystyle{IEEEtran}
\bibliography{bibtex}

\end{document}